\documentclass[journal]{IEEEtran}
\usepackage[backend=biber]{biblatex}
\usepackage{arydshln}
\usepackage{graphicx}
\usepackage{amsmath}
\usepackage{amssymb}
\usepackage[table]{xcolor}
\usepackage{pifont}
\usepackage{multirow}
\usepackage[T5]{fontenc}
\usepackage{diagbox}
\usepackage[symbol]{footmisc}

\begin{document}
\title{UIT-OpenViIC: A Novel Benchmark for Evaluating Image Captioning in Vietnamese}

\author{Doanh C. Bui \qquad Nghia Hieu Nguyen \qquad Khang Nguyen \footnote{*Corresponding author} \\
University of Information Technology, Ho Chi Minh city, Vietnam\\
Vietnam National University, Ho Chi Minh City, Vietnam\\
{\tt\small 19521366@gm.uit.edu.vn, 19520178@gm.uit.edu.vn, khangnttm@uit.edu.vn}
}
\maketitle

\begin{abstract}
Image Captioning is one of the vision-language tasks that still interest the research community worldwide in the 2020s. MS-COCO Caption benchmark is commonly used to evaluate the performance of advanced captioning models, although it was published in 2015. Recent captioning models trained on the MS-COCO Caption dataset only have good performance in language patterns of English; they do not have such good performance in contexts captured in Vietnam or fluently caption images using Vietnamese. To contribute to the low-resources research community as in Vietnam, we introduce a novel image captioning dataset in Vietnamese, the \textbf{Open}-domain \textbf{Vi}etnamese \textbf{I}mage \textbf{C}aptioning dataset (UIT-OpenViIC). The introduced dataset includes complex scenes captured in Vietnam and manually annotated by Vietnamese under strict rules and supervision. In this paper, we present in more detail the dataset creation process. From preliminary analysis, we show that our dataset is challenging to recent state-of-the-art (SOTA) Transformer-based baselines, which performed well on the MS COCO dataset. Then, the modest results prove that UIT-OpenViIC has room to grow, which can be one of the standard benchmarks in Vietnamese for the research community to evaluate their captioning models. Furthermore, we present a CAMO approach that effectively enhances the image representation ability by a multi-level encoder output fusion mechanism, which helps improve the quality of generated captions compared to previous captioning models. The dataset is available online for non-commercial research at \url{https://uit-together.github.io/datasets/UIT-OpenViIC/}.
\end{abstract}

\IEEEpeerreviewmaketitle

\section{Introduction}
Vision-language task has attracted recent study because of its challenge but potential realistic application. One of the most interesting vision-language task is image captioning (IC), which is the task of generating the description for images.  This ability of machines has a wide range of applications where the most popular ones are the content-based image retrieval problem which is the premise of other crucial fields: biomedicine, commerce, education, digital libraries as well as web searching \cite{hossain2019survey}. Lots of datasets have been published for evaluating the solution for image captioning. The most commons can be listed are MS-COCO Caption \cite{chen2015microsoft} and Flickr30k \cite{flickr30K}, which were annotated via large community annotators and have been used widely. Inspired by these datasets, many other datasets were proposed to tackle other sub-directions of image captioning, e.g., GoodNews \cite{biten2019good}, TextCaps \cite{textcaps}, CrowdCaption \cite{crowdedcaption}, and VisualGenome \cite{krishna2017visual}. However, all above datasets were annotated in English, which conducts a barrier for other research communities using different languages such as Vietnamese which is one of the low-resource languages. Nonetheless, there have not been many studies that thoroughly explore the challenges that exist in Image Captioning in the Vietnamese language. To this end, we conducted a novel and first large-scale open-domain dataset for researching image captioning in Vietnamese. Main contributions in our study are listed as follow:

\begin{figure*}[!http]
\centerline{\includegraphics[width=17cm]{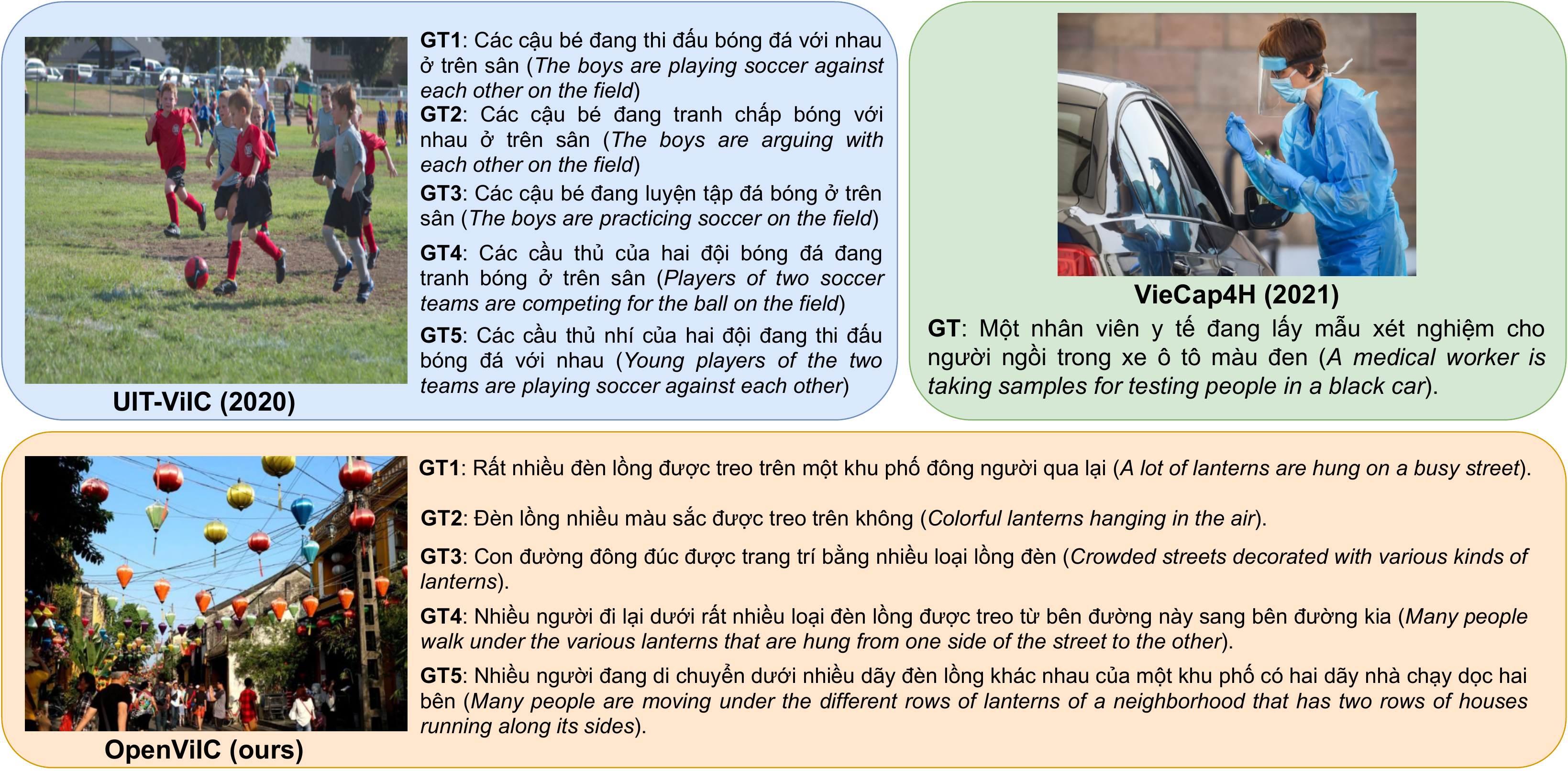}}
\caption{Qualitative comparison between UIT-OpenViIC (ours) and two previous Vietnamese dataset for evaluating image captioning: UIT-ViIC \cite{uitviic} and VieCap4H \cite{le2021vlsp}.}
\label{quanlitative-comparison}
\end{figure*}

\begin{itemize}
    \item We introduce the first open-domain Vietnamese dataset named UIT-OpenViIC. To the best of our knowledge, our proposed dataset is the largest manually annotated by Vietnamese annotators under our supervision. Moreover, the UIT-OpenViIC dataset is a unique dataset, including images having contexts relevant to daily life and culture in Vietnam. Our annotators are required to provide free-form captions for each image they are given, which leads our dataset to become diverse and challenging to the vision-language research communities in Vietnam, particularly the vision-language research community over the world.
    \item We evaluate recent SOTA methods proposed for image captioning task on the MS COCO dataset including ORT \cite{ort}, \(\mathcal{M}^2\) Transformer \cite{m2transformer}, RSTNet \cite{rstnet}, DLCT \cite{luo2021dual}, DIFNet \cite{difnet}, and MDSANet \cite{mdsanet}. Experimental results show that our dataset challenges these SOTA methods in both vision aspect and linguistic aspect, motivates the vision-language research community to develop the captioning approaches for better performance in Vietnamese.
    \item We present the CAMO approach, which stands for Cross-Attention on Multi-level Outputs. Our proposed CAMO module is used to effectively multi-level outputs from encoder layers via a self-attention mechanism, which can significantly boost the captioning models' performance.
\end{itemize}

The rest of this paper is structured as follows: Section \ref{sec:relatedwork} mentions the most common English datasets and current Vietnamese datasets proposed to research the image captioning task. Section \ref{sec:UIT-OpenViICdataset} describes clearly the data creation process of our UIT-OpenViIC dataset. Section \ref{sec:proposedmethod} presents our approach that performs better than other captioning models in our UIT-OpenViIC dataset. Section \ref{sec:experimentalre} reports the series of experiments. Section \ref{sec:conclusion} summarizes our study succinctly and provides the research community problems that need to grow.

\section{Related works}
\label{sec:relatedwork}

\begin{table*}[http]
\centering
\caption{The boldface \textbf{\textcolor{red}{red}} text indicates the highest number among all mentioned datasets. The boldface \textbf{\textcolor{blue}{blue}} text indicates that number among Vietnamese datasets. * denotes datasets whose statistics are only obtained from training \& validation sets (or may have public-test set) because the private-test set is not publicly published.}
\resizebox{1\textwidth}{!}{\begin{tabular}{lccccccc}
\hline
\multicolumn{1}{c}{\textbf{Dataset}} & \textbf{Images} & \vtop{\hbox{\strut \textbf{Captions}}\hbox{\strut \textbf{per image}}} & \vtop{\hbox{\strut \textbf{Average}}\hbox{\strut \textbf{length}}} & \vtop{\hbox{\strut \textbf{Objects}}\hbox{\strut \textbf{per image}}} & \vtop{\hbox{\strut \textbf{\#No}}\hbox{\strut \textbf{of objects}}} & \vtop{\hbox{\strut \textbf{\#No}}\hbox{\strut \textbf{of verbs}}} & {\textbf{Contexts}} \\
\hline
\multicolumn{8}{c}{\textit{English Datasets}}                                                                                                                                                     \\ \hline
MS-COCO Caption (2014) \cite{chen2015microsoft} \footnote[1]             & \textbf{\textcolor{red}{123,287}}         & 5                       & 10                      & 10.53                        & 19,665                          & 8,478 & Open Domain                        \\
Flickr30k (2015) \cite{flickr30K}                     & 31,783          & 5                       & 12                      & 12.09                        & 44,518                      & 5,777 & Open Domain                        \\
VisualGenome (2016) \cite{krishna2017visual}                  & 108,077         & 1 for 1 region          & 5                       & \textbf{\textcolor{red}{39.94}}                        & \textbf{\textcolor{red}{47,127}}                     & \textbf{\textcolor{red}{17,652}} & Open Domain                        \\
TextCaps (2020) \cite{textcaps} \footnote[1]                   & 28,408          & \textbf{\textcolor{red}{5.1}}                     & 12                      & 11.21                          & 20,365                           & 6,399 & Scene texts                        \\
CrowdCaption (2022) \cite{crowdedcaption}                  & 11,161          & 4.4                     & \textbf{\textcolor{red}{20}}                      & 1.50                       & 164                           & 766 & Person, Crowded scenes                        \\ \hline
\multicolumn{8}{c}{\textit{Vietnamese Datasets}}                                                                                                                                                           \\ \hline
UIT-ViIC (2020) \cite{uitviic}                      & 3,850           & 5                       & 11 to 15                & 10.42                      & 805                         & 588 & Ball sports                      \\
VieCap4H (2021) \cite{le2021vlsp} \footnote{}                     & 10,068          & 1.15                    & 11.89                   & 4.8                       & 1,228                       & 768 & Healthcare                      
\\ \hline \hline
\cellcolor[HTML]{F3F3F3}\textbf{UIT-OpenViIC (ours)}             & \cellcolor[HTML]{F3F3F3}\textbf{\textcolor{blue}{13,100}} & \cellcolor[HTML]{F3F3F3}\textbf{\textcolor{blue}{4.7}}                     & \cellcolor[HTML]{F3F3F3}\textbf{\textcolor{blue}{17.61}}                   & \cellcolor[HTML]{F3F3F3}\textbf{\textcolor{blue}{20.5}}             & \cellcolor[HTML]{F3F3F3}\textbf{\textcolor{blue}{4,177}}              & \cellcolor[HTML]{F3F3F3}\textbf{\textcolor{blue}{2,425}} & \cellcolor[HTML]{F3F3F3}Open Domain \\
\hline
\end{tabular}}
\label{tab:stastitic+comparison}
\end{table*}

\subsection{Common Benchmark Datasets}

\textbf{Flickr30K.} Flickr30k dataset and its precedent Flickr8k dataset are the one of the first datasets proposed to evaluate image captioning task. Flickr-family datasets consist of images about daily activities, scenes and events (8,092 images in Flickr8k and 31,783 in Flickr30k) harvested from Flickr platform, with its associated captions achieved by croudsourcing. For each image in the Flickr30k dataset, annotators are required to annotate at least five captions. After annotating the whole images, captions were refined and preprocessed by using some rule templates \cite{flickr30K} such as normalizing captions, replacing nouns by hypernyms and extract simpler constituents.

\textbf{MS-COCO Caption.} For evaluating image captioning tasks, many benchmark datasets are widely used. The most common dataset is MS-COCO Caption \cite{chen2015microsoft}, which includes over 120K images for training. Each image is annotated by five independent annotators using the AMT interface with seven strict rules. The dataset is split into training and validation. The online evaluation is done by submitting the results on provided evaluation server. However, many studies \cite{m2transformer, aoanet, rstnet, luo2021dual, difnet, mdsanet, anderson2018bottom} also did the offline evaluation of their proposed captioning models on the ``Karpathy" test split, \cite{7298932}, which is a part of the validation set in the MS-COCO Caption dataset.

\textbf{VisualGenome.} \citeauthor{krishna2017visual} \cite{krishna2017visual} aimed to research how to model the ability to identify the objects and their relationship and then correctly answer the given question. Visual Genome was used as a resource to train the Faster-RCNN method \cite{fasterrcnn} for the Image Captioning task when Sherdade \citeauthor{ort} \cite{ort} proposed their approach that uses box coordinates of detected objects in order to enhance the self-attention of transformer \cite{transformer}.

\textbf{TextCaps.} Most common datasets concentrate on exploring visual information of objects and their relationships in images. On the other hand, \citeauthor{textcaps} \cite{textcaps} argued that scene texts in images also play a crucial role when providing visual information for a method of learning and captioning the image. Therefore, to prepare a resource for the research community to explore and develop approaches that have the ability to use the information of scene text together with objects in images, Sidorow \citeauthor{textcaps} released the first novel dataset, the TextCaps dataset, to tackle the image captioning task with reading comprehension. 

\textbf{CrowdCaption.} \citeauthor{crowdedcaption} \cite{crowdedcaption} criticized that although the image captioning task attracts the most interest from researchers from all over the world, captioning for crowded scenes is rarely explored due to the shortage of relevant datasets. To motivate the research community to bridge the gap of captioning a crowded scene in an image captioning task, Wang \citeauthor{crowdedcaption} \cite{crowdedcaption} constructed a novel and high-quality dataset, the CrowCaption dataset.

\begin{figure*}[ht]
\centerline{\includegraphics[width=15cm]{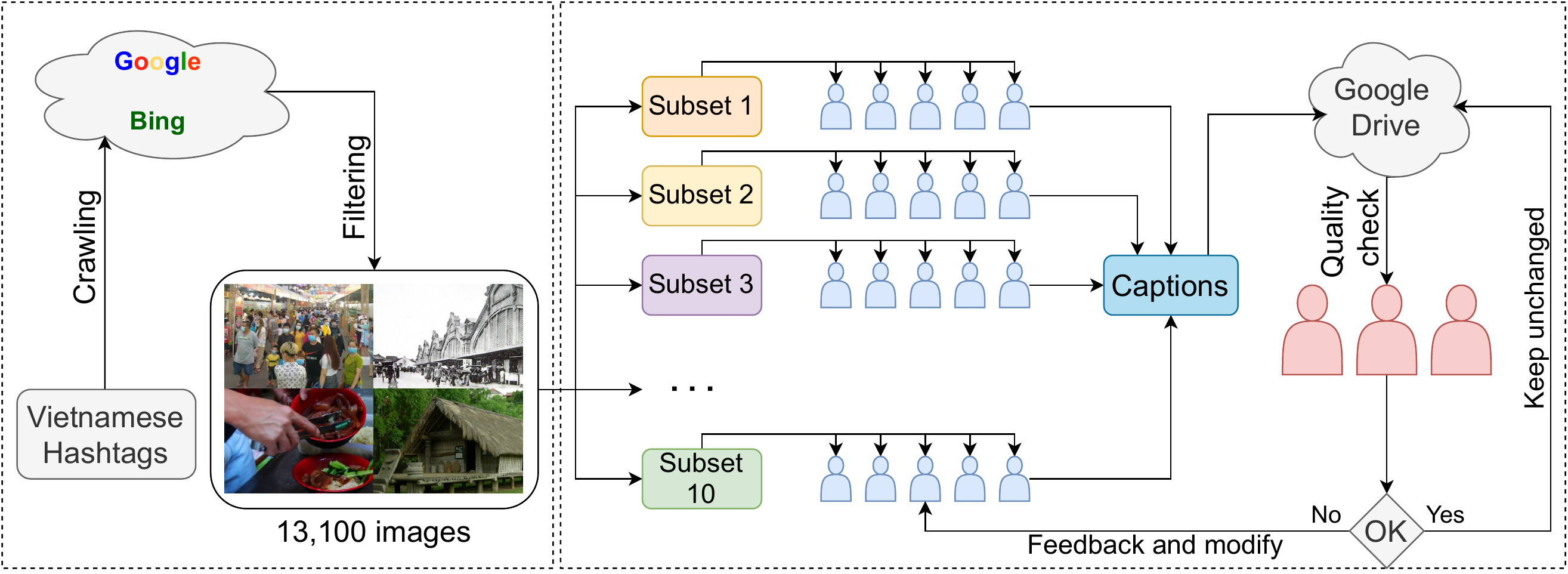}}
\caption{The pipeline of creating our UIT-OpenViIC dataset. A thousand images are first collected from Google and Bing search engines, then filtered, and 13,100 images are selected in the final. The raw dataset is split into ten subsets, and each subset is assigned to one student in a group. Students in the same group are given the same subgroup but do not know each other to keep quality.}
\label{dataset-pipeline}
\end{figure*}

\subsection{Vietnamese Image Captioning Datasets}

\textbf{UIT-ViIC.} \citeauthor{uitviic} \cite{uitviic} released the first dataset, the UIT-ViIC dataset, for researching image captioning in Vietnamese. This dataset consists of 3,850 images relevant to sports played with balls collected from MS COCO dataset. For each image, annotators were asked to give at least 5 captions, leads to 19,250 captions in total of the UIT-ViIC dataset.

\textbf{VieCap4H.} \citeauthor{le2021vlsp} \cite{le2021vlsp} released the second dataset, the VieCap4H dataset, for image captioning in Vietnamese under the shared task challenge held by the Vietnamese Language and Speech Processing (VLSP). VieCap4H dataset consists of 10,068 images collected from articles in healthcare domain with its associated 11,563 captions. Together with UIT-ViIC \cite{uitviic}, the this dataset is the second specific-domain image captioning dataset in Vietnamese.

\section{The UIT-OpenViIC dataset}
\label{sec:UIT-OpenViICdataset}

\subsection{Dataset Creation Process}

First, we crawled the images mainly from two sources: Google and Bing with Vietnamese keywords. After that, we filtered some coincident images that did not contain enough information to be described. The final number of selected images is 13,100 images. Then, we split them into ten subsets; each subset included 1,310 images. For annotation, we have a team including 51 individual students at the Faculty of Software Engineering, University of Information Technology. These 51 students are split into ten groups, each contains five students. Each student in a group was assigned one subset to annotate. All members in a group were given the same subset to obtain approximately five captions per image. Notably, we did not let the students in a group know each other, and the file name of images was also set as random strings to ensure they did not copy each other. The creation pipeline is illustrated in Figure \ref{dataset-pipeline}. Inspired by the rules in the MS-COCO Caption creation pipeline \cite{chen2015microsoft}, we also asked the students follow strictly seven rules to keep the high quality of annotated captions:

\textbf{Each sentence must have at least ten words.} We want the generated captions to become diverse, hence we asked the annotators to provide at least ten words per caption. 

\textbf{Annotators should clearly describe visual information that appears in images.} We ask the annotators try to describe all visual information that appear in images such as objects, interaction between objects, activities, etc. 

\textbf{Common English words can be used: TV, laptop, etc.} In Vietnamese or other languages, some common English words are still widely understood by a large number of people because of being used with a high frequency. Therefore, we encourage the annotators to keep these common English words to describe objects if needed.

\textbf{DO NOT describe future events.} We suppose that future events do not directly appear in the scene and only can be inferred by the human via available objects, actions, and current contexts. Future circumstances may confuse the model because future things do not appear in the scene and can not contribute to the pattern recognition process.

\textbf{DO NOT describe the scene texts.} Because describing the scene texts need the multimodal-based model, which receives not only the input RGB image but also the features of regions of scene texts and text embedding vectors of recognized texts. Therefore, we did not encourage the annotators to mention text information in the captions.

\textbf{DO NOT put personal emotion in the captions. The annotated captions must be objective.} Because personal emotion may affect the thinking of images' context, we encourage the annotators not to put their personal feelings about events, objects, or actions that occur in images. All captions are only general visual descriptions, as much as possible.

We also designed the GUI interface for annotators \footnote[2]{\url{https://github.com/caodoanh2001/UIT-OpenViIC-labeller}}, which shows the image on the left and only one text box on the right. At any time of annotating 1,310 images, the annotators can safely back up the captions in Google Drive via the ``Backup" button. A team of senior researchers will carefully recheck the quality of captions and require the annotators to fix some mistakes, which were almost spelling errors.

\subsection{Statistics}

In total, the UIT-OpenViIC dataset has 13,100 images with 61,241 appropriate captions. To split the dataset into train-dev-test partitions, we took approximately 30\% of total images and all its captions for the validation set and the test set, where each set has approximately 15\% of total images and left the remaining images as the training set. We randomly selected images for the validation set and test set using the uniform distribution so that all images have the same chance to participate in the dev set and test set. The sampling process for the validation set (test set, respectively) was performed until the ratio of selected images was equal to or higher than 15\% (15\%, respectively). Statistically, the training set of the UIT-OpenViIC has 9,088 images with 41,238 captions; the validation set has 2,011 images with 10,002 captions, and the test set has 2,001 images with 10,001 captions.

\begin{figure}[ht]
\centerline{\includegraphics[width=9cm]{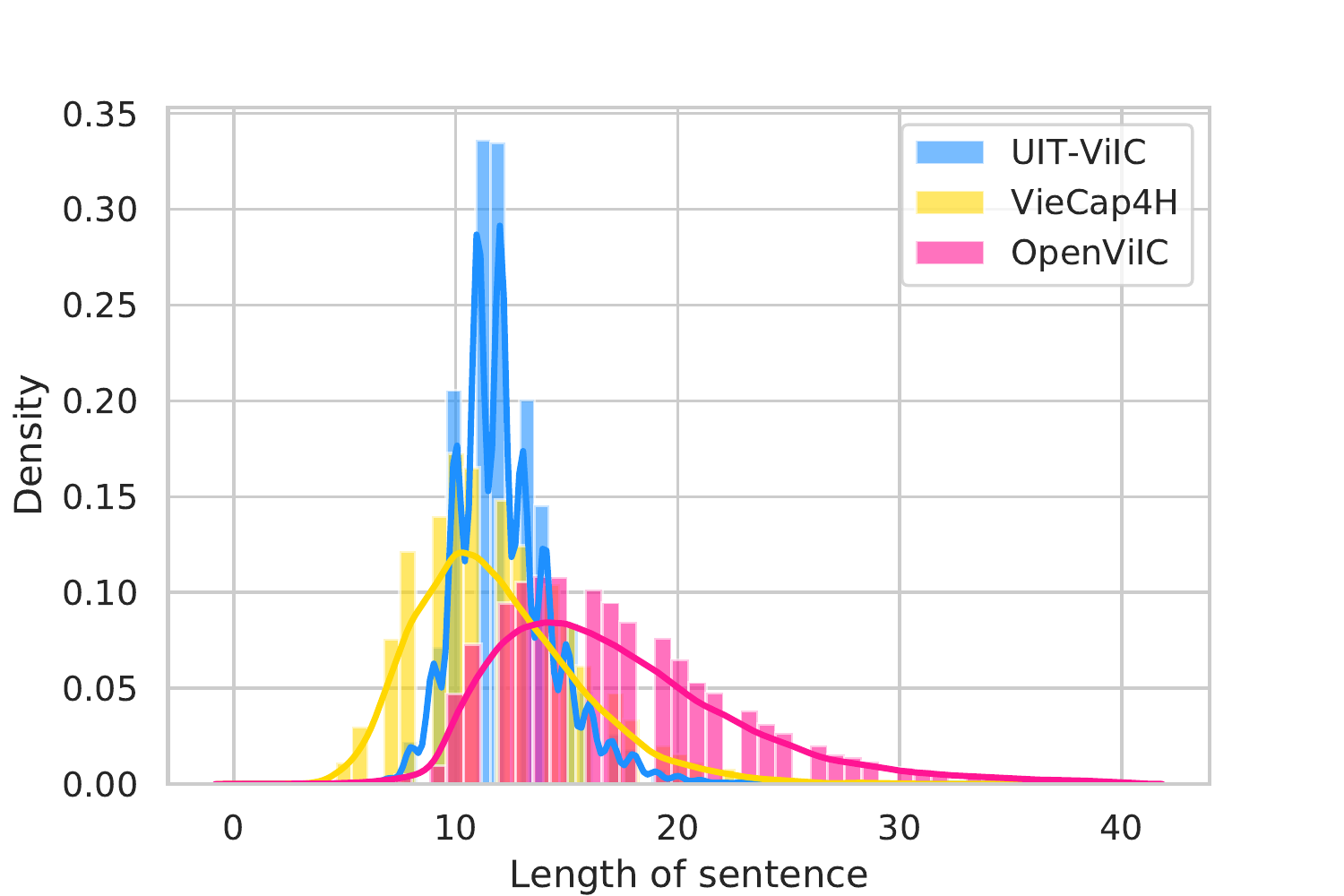}}
\caption{The comparison of distribution of length of sentence between three Vietnamese captioning datasets.}
\label{distribution-sentence-length}
\end{figure}

In Table \ref{tab:stastitic+comparison}, we provide the statistics of standard datasets, which generated captions in English, and our Vietnamese UIT-OpenViIC dataset. The characteristics we want to compare between datasets are \textbf{the number of images}, \textbf{the ratio of captions per image}, \textbf{the average length}, and \textbf{the number of objects and verbs}. To calculate number of objects and verbs, we use a pre-trained Vietnamese POS tagger \cite{vu2018vncorenlp}. Tokens are classified as ``Nouns'' and are considered as ``Objects.'' Then, we only count the unique nouns and get the final number of objects. A similar process is done with the number of verbs. As results, our UIT-OpenViIC outperforms all Vietnamese datasets regarding the number of images and captions and the number of objects and verbs. It proves that our dataset is much more diverse and describes more detailed events in a single image. We apply the same analysis strategy with English benchmarks but use the pre-trained spaCy POS Tagger \footnote[3]{\url{https://spacy.io/}}. 

In general, our UIT-OpenViIC dataset has the highest number of images and captions among the Vietnamese datasets. UIT-ViIC \cite{uitviic} can be considered the first Vietnamese dataset for image captioning. However, the images were obtained only from the MS-COCO Caption dataset with handball sports contexts. Although the ratio of captions per image is the highest among Vietnamese datasets, the number of image-caption pairs is still limited because they have only 3,850 images. In spite of having a larger number of images, VieCap4H \cite{le2021vlsp} only focused on healthcare contexts, and the ratio of captions per image was meager (1.15). 
Compared to analysis results on standard datasets in English, our dataset has more objects per image than MS-COCO Caption \cite{chen2015microsoft}, Flickr30K \cite{flickr30K} and TextCaps \cite{textcaps}. Our number of captions per image is also comparable to most English datasets.

\begin{figure*}[http]
\centerline{\includegraphics[width=17cm]{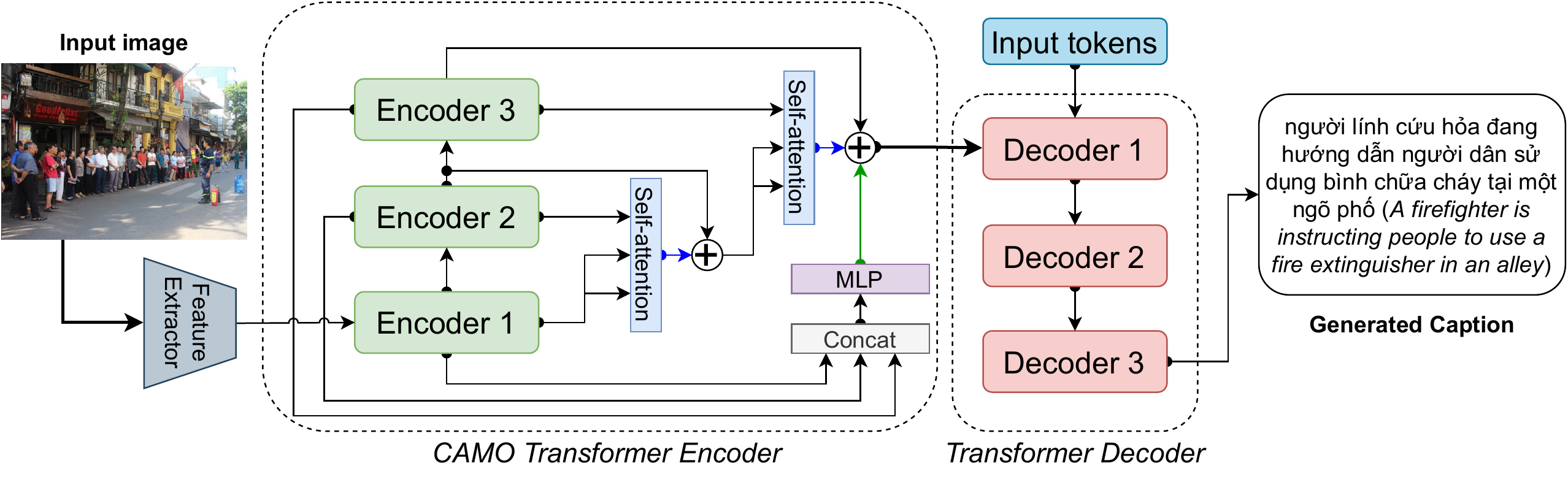}}
\caption{The illustration of the proposed CAMO scheme for multi-level encoder outputs fusion.}
\label{camo}
\end{figure*}

\subsection{Qualitative Comparison}
We also provide the qualitative comparison between three Vietnamese datasets in Figure \ref{quanlitative-comparison}. It can be observed that annotated captions in our UIT-OpenViIC dataset are more diverse. Five captions for an image in the UIT-ViIC dataset have high similarity, and the frequency of each word is high, such as ``Các cậu bé (The boys)", ``trên sân (on the field),'' ``bóng đá (football).'' The reason is that only one person annotates five captions in UIT-ViIC; brainstorming and describing five times for one picture may make the diversity and quality of captions low. In the VieCap4H dataset, each image is annotated with only one to three captions; therefore, captioning models can not learn distinct aspects of images well. Our UIT-ViIC dataset also has approximately five captions per image, but they are much more varied. Each word does not appear many times, and each caption describes different aspects of the input image. The distributions of sentence length of UIT-ViIC, VieCap4H and our UIT-OpenViIC are illustrated in Figure \ref{distribution-sentence-length}. Lengths of captions in UIT-ViIC dataset tend to be denser at approximately 10 to 15 tokens. VieCap4H has better distribution in the same range of tokens. The description sentence length distribution in our UIT-OpenViIC dataset is much more even and more excellent.

\section{CAMO: Cross-Attention on Multi-level Encoder Outputs}
\label{sec:proposedmethod}

To enhance the quality of generated captions in our UIT-OpenViIC dataset using Transformer-based models, we propose \textbf{C}ross-\textbf{A}ttention on \textbf{M}ulti-level Encoder \textbf{O}utputs (CAMO) module, which is used to augment the encoded outputs from multi-level encoder layers. In the original Transformer architecture \cite{transformer}, there are \(N\) encoder layers; each layer takes the outputs from previous layers as inputs and continues to learn the high-level latent spaces. Commonly, only the final output from the last encoder layer is taken to the Transformer Decoder, this may ignore some valuable information from the low-level encoder layers, and the Transformer Decoder does not receive the best representation from encoder layers. Therefore, we propose the CAMO module, which effectively fuses all low to high-level encoder outputs, to enhance the ability of encoded outputs during the decoding scheme. 

First, given \(N\) encoder layers, we first calculate the high-level representations from encoder layers, which are formulated as Equation \ref{eq:0}:

\begin{equation}
    Z_i = 
    \begin{cases}
      EncoderLayer_i(V), & \text{if } l = 0 \\
      EncoderLayer_i(Z_{i-1}), & \text{otherwise}
    \end{cases},
\label{eq:0}
\end{equation}
where $Z_i \in \mathbb{R}^{d_{model}}$ is the high-level representation computed from $i^{th}$ encoder layer. When $i = 0$, the encoder layer takes visual features $V$ of images as inputs.

First, we want to learn the joint representation from multi-level encoder outputs to take advantage of all valuable information from them. The proposed scheme to learn these is simple but effective, which can be formulated as Equations \ref{eq:1}, \ref{eq:2}:

\begin{equation}
Z_{f} = \text{Concatenate}(Z_1, Z_2, ..., Z_N),
\label{eq:1}
\end{equation}

\begin{equation}
Z_{f} = \text{LeakyRELU}(\text{MLP}(Z_{f})),
\label{eq:2}
\end{equation}
where $\text{MLP}$ is the projection that learn to transform $Z_f \in \mathbb{R}^{d_{model \times 3}}$ to $Z_f \in \mathbb{R}^{d_{model}}$; $\text{LeakyRELU}$ is the activation that is used to scale the range of $Z_f$.
 
After that, we use the cross self-attention scheme to compute the attention weights of encoder output from $i^{th}$ encoder layer using encoder output from $(i-1)^{th}$ encoder layer. These can be formulated as Equations \ref{eq:3}, \ref{eq:4}:

\begin{equation}
Q = W^Q \times Z_{i}; K = W^K \times Z_{i-1}; V = W^V \times Z_{i-1},
\label{eq:3}
\end{equation}

\begin{equation}
Z_i = \alpha \times softmax ( \frac{Q \times K^T}{ \sqrt{d_k} } ) V + Z_i,
\label{eq:4}
\end{equation}
where $W^Q$, $W^K$, and $W^V$ are learned weight matrices.
The $Z_i$ is now calculated as the attention weights again using encoder output $Z_{i-1}$. It helps $Z_i$ take attention knowledge from the previous encoder layer and augment the representation ability. However, we do not want to affect too much what we have just learned from the current $i^{th}$ encoder layer. Therefore, the hyperparameter $\alpha$ is used to control the contribution of the output from the cross-self-attention scheme. We also conduct the ablation study to explore which $\alpha$ is the best.

Finally, the skip connection is used to fuse $Z_N$ and $Z_f$ to augment the knowledge from joint representation to the final encoder outputs $Z_N$, which is formulated as Equation \ref{eq:5}:

\begin{equation}
Z_o = \beta \times Z_f + Z_N,
\label{eq:5}
\end{equation}
where \(\beta\) is the hyperparameter to control the contribution of joint learned joint representation to the final encoded output.
The final representation \(Z_o\) now effectively presents the encoded output from Transformer Encoder, and we fit it into the Transformer Decoder to learn to generate words in a caption.
By default, we set \(\alpha = 0.1\), and \(\beta = 0.2\).

\begin{table*}[http]
\centering
\caption{The experimental results of recent state-of-the-art models on test split of UIT-OpenViIC dataset. The \textbf{best} performance is marked in boldface.}
\resizebox{\textwidth}{!}{
\begin{tabular}{cccccccccc}
\hline
\multirow{2}{*}{\textbf{Methods}} & \multirow{2}{*}{\textbf{Features}} & \textbf{BLEU@1} & \textbf{BLEU@2} & \textbf{BLEU@3} & \textbf{BLEU@4} & \textbf{METEOR} & \textbf{ROUGE} & \multicolumn{2}{c}{\textbf{CIDEr}}  \\
\cline{3-8} \cline{9-10}
 &  & \multicolumn{6}{c}{\textbf{UIT-OpenViIC}} & \textbf{UIT-OpenViIC} & \textbf{MS-COCO} \\
 \hline
Transformer \cite{transformer} & \(\mathcal{R}\) & 76.7125 & 61.2546 & 47.3191 & 36.2317 & 30.2674 & 47.7332 & 66.9046 & - \\
ORT \cite{ort} & \(\mathcal{R}\) & 76.9279 & 60.9233 & 46.7828 & 35.5641 & 30.4797 & 47.6364 & 66.888 & 128.3  \\
\(\mathcal{M}^2\) Transformer \cite{m2transformer} & \(\mathcal{R}\) & 76.0137 & 60.7039  & 47.2007 & 36.0695 & 30.1968 & 48.3066 & 68.7317 & 131.2 \\
AoANet \cite{aoanet} & \(\mathcal{R}\) & 77.3173 & 61.5353 & 47.7429 & 36.6519 & 30.7232 & 48.3634 & 69.4182 & 129.8 \\ 
RSTNet \cite{rstnet} & \(\mathcal{G}\) + \(\mathcal{L}\) & 77.3713 & 61.7489 & 48.0342 & 36.7744 & 30.7143 & 48.9556 & 72.7171 & 133.3 \\
DLCT \cite{luo2021dual} & \(\mathcal{R}\) + \(\mathcal{G}\) & 77.3665 & 61.9476 & 48.3324 & 37.2237 & 30.8583 & 49.1286 & 72.6897 & 133.5 \\
DIFNet \cite{difnet} & \(\mathcal{G}\) + \(\mathcal{S}\) & 77.6909 & 62.4044 & 48.9048 & 37.6666 & 30.8961 & 49.1255 & 74.1907 &  \textbf{136.2} \\
MDSANet \cite{mdsanet} & \(\mathcal{G}\) & \textbf{78.1167} & \textbf{63.0966} & \textbf{49.8570}  & \textbf{38.7951} & \textbf{31.0460}  & \textbf{49.7740}  & \textbf{75.5796} & 135.1 \\
\hline
\end{tabular}}
\label{tab:expresults}
\end{table*}

\section{Experimental Results}
\label{sec:experimentalre}

\subsection{Baseline Methods}
\label{sec:baseline}

We choose eight Transformer-based state-of-the-art models for image captioning to evaluate their performance on our UIT-OpenViIC dataset: Transformer \cite{transformer}, ORT \cite{ort}, \(\mathcal{M}^2\) Transformer \cite{m2transformer}, RSTNet \cite{rstnet}, DLCT \cite{luo2021dual}, DIFNet \cite{difnet}, MDSANet \cite{mdsanet}. Transformer \cite{transformer} is a standard Transformer architecture whose idea is proposed by \citeauthor{transformer} \cite{transformer}. ORT was the first study that used the box coordinates as auxiliary information, better guiding the self-attention mechanism. \(\mathcal{M}^2\)  implements the idea of using all encoder outputs and fitting them into the decoder instead of using only features from the last encoder layer. RSTNet \cite{rstnet} is a transformer-based model with two main components: Grid-augmented (GA) and Adaptive Attention (AA). GA was proposed to add the attention weights to the relation features between grids to guide the self-attention mechanism effectively. Adaptive Attention is used to concatenate the language signals obtained from the pre-trained language model with encoder outputs, which worked better with non-visual words. DLCT \cite{luo2021dual} encodes both region features and grid features via self-attention and cross-self-attention mechanisms to give the primary captioning model awareness of the input image's local and global texture features. DIFNet \cite{difnet} model is a captioning model based on Transformer architecture that uses multi-representation for input images. In detail, \citeauthor{difnet} \cite{difnet} extract visual features of images from a CNN-based model. Simultaneously, a segmentation model also obtains the segmentation mask of the input image. Then, they use the fusion mechanism, similar to vanilla self-attention \cite{nagrani2021attention}, to fuse the segmentation mask and visual features. MDSANet \cite{mdsanet} treats the bounding boxes as positional information to measure the distances between objects in images. Furthermore, they consider a multi-head attention layer as a branch and repeat the self-attention mechanism in a loop to learn more valuable information from visual features.

\begin{table}[http]
\centering
\caption{The experimental results of using CAMO (\(\alpha = 0.1\), \(\beta = 0.2\)) (\(\mathcal{C}\)) on existing captioning models. The \textbf{best} performance is marked in boldface.}
\resizebox{0.48\textwidth}{!}{\begin{tabular}{lccccc}
\hline
\textbf{Methods}      & \textbf{BLEU@1} & \textbf{BLEU@4} & \textbf{METEOR} & \textbf{ROUGE} & \textbf{CIDEr} \\
\hline
Transformer           & 76.7125      & 36.2317      & 30.2674         & 47.7332        & 66.9046        \\ \cdashline{1-6} 
\multirow{2}{*}{Transformer + \(\mathcal{C}\)}    & \cellcolor[HTML]{F3F3F3}{77.7135}      & \cellcolor[HTML]{F3F3F3}{37.7876}      & \cellcolor[HTML]{F3F3F3}{30.7159}         & \cellcolor[HTML]{F3F3F3}{48.7078}        & \cellcolor[HTML]{F3F3F3}{71.8213}        \\
    & \cellcolor[HTML]{F3F3F3}{(+1.001)}      & \cellcolor[HTML]{F3F3F3}{(+1.5559)}      & \cellcolor[HTML]{F3F3F3}{(+0.4485)}         & \cellcolor[HTML]{F3F3F3}{(+0.9746)}        & \cellcolor[HTML]{F3F3F3}{(+4.9167)}        \\
\hline
ORT           & 76.9279      & 35.5641      & 30.4797         & 47.6364        & 66.8880        \\ \cdashline{1-6} 
\multirow{2}{*}{ORT + \(\mathcal{C}\)}    & \cellcolor[HTML]{F3F3F3}{75.7919}      & \cellcolor[HTML]{F3F3F3}{36.5132}      & \cellcolor[HTML]{F3F3F3}{30.1419}         & \cellcolor[HTML]{F3F3F3}{48.0625}        & \cellcolor[HTML]{F3F3F3}{69.3468}        \\
    & \cellcolor[HTML]{F3F3F3}{(-1.136)}      & \cellcolor[HTML]{F3F3F3}{(+0.9491)}      & \cellcolor[HTML]{F3F3F3}{(-0.3378)}         & \cellcolor[HTML]{F3F3F3}{(+0.4261)}        & \cellcolor[HTML]{F3F3F3}{(+2.4588)}        \\
\hline
AoANet           & 77.3173      & 36.6519      & 30.7232         & 48.3634        & 69.4182        \\ \cdashline{1-6} 
\multirow{2}{*}{AoANet + \(\mathcal{C}\)}    & \cellcolor[HTML]{F3F3F3}{77.5228}      & \cellcolor[HTML]{F3F3F3}{38.1818}      & \cellcolor[HTML]{F3F3F3}{31.0070}         & \cellcolor[HTML]{F3F3F3}{49.2816}        & \cellcolor[HTML]{F3F3F3}{73.9883}        \\
    & \cellcolor[HTML]{F3F3F3}{(+0.2055)}      & \cellcolor[HTML]{F3F3F3}{(+1.5299)}      & \cellcolor[HTML]{F3F3F3}{(+0.2838)}         & \cellcolor[HTML]{F3F3F3}{(+0.9182)}        & \cellcolor[HTML]{F3F3F3}{(+4.5701)}        \\
\hline
\(\mathcal{M}^2\) Transformer        & 76.0317      & 36.0695      & 30.1968         & 48.3066        & 68.7317        \\ \cdashline{1-6}
\multirow{2}{*}{\(\mathcal{M}^2\) Transformer + \(\mathcal{C}\)} & \cellcolor[HTML]{F3F3F3}{78.1709}      & \cellcolor[HTML]{F3F3F3}{38.3018}      & \cellcolor[HTML]{F3F3F3}\textbf{{31.2974}}         & \cellcolor[HTML]{F3F3F3}{49.1784}        & \cellcolor[HTML]{F3F3F3}{71.7504}        \\
 & \cellcolor[HTML]{F3F3F3}{(+2.1392)}      & \cellcolor[HTML]{F3F3F3}{(+2.2323)}      & \cellcolor[HTML]{F3F3F3}\textbf{{(+1.1006)}}         & \cellcolor[HTML]{F3F3F3}{(+0.8718)}        & \cellcolor[HTML]{F3F3F3}{(+3.0187)}        \\
\hline
RSTNet                & 77.3713      & 36.7744      & 30.7143         & 48.9556        & 72.7171        \\ \cdashline{1-6}
\multirow{2}{*}{RSTNet + \(\mathcal{C}\)}         & \cellcolor[HTML]{F3F3F3}{77.7370}      & \cellcolor[HTML]{F3F3F3}{38.0970}      & \cellcolor[HTML]{F3F3F3}{31.0655}         & \cellcolor[HTML]{F3F3F3}{49.3731}        & \cellcolor[HTML]{F3F3F3}{74.1869}        \\
         & \cellcolor[HTML]{F3F3F3}{(+0.3657)}      & \cellcolor[HTML]{F3F3F3}{(+1.3226)}      & \cellcolor[HTML]{F3F3F3}{(+0.3512)}         & \cellcolor[HTML]{F3F3F3}{(+0.4175)}        & \cellcolor[HTML]{F3F3F3}{(+1.4698)}        \\
\hline
DIFNet                & 77.6909      & 37.6666      & 30.8961         & 49.1255        & {74.1907}        \\ \cdashline{1-6}
\multirow{2}{*}{DIFNet + \(\mathcal{C}\)}         & \cellcolor[HTML]{F3F3F3}{77.7191}      & \cellcolor[HTML]{F3F3F3}{38.0533}      & \cellcolor[HTML]{F3F3F3}{31.0573}         & \cellcolor[HTML]{F3F3F3}{49.2749}        & \cellcolor[HTML]{F3F3F3}73.1631        \\
       & \cellcolor[HTML]{F3F3F3}{(+0.0282)}      & \cellcolor[HTML]{F3F3F3}{(+0.3867)}      & \cellcolor[HTML]{F3F3F3}{(+0.1612)}         & \cellcolor[HTML]{F3F3F3}{(+0.1494)}        & \cellcolor[HTML]{F3F3F3}(-1.0276)        \\
\hline
MDSANet               & 78.1167      &  38.7951      & 31.0460         & 49.7740        & 75.5796        \\ \cdashline{1-6}
\multirow{2}{*}{MDSANet + \(\mathcal{C}\)}        & \cellcolor[HTML]{F3F3F3}\textbf{{78.7065}}      & \cellcolor[HTML]{F3F3F3}\textbf{{38.8552}}      & \cellcolor[HTML]{F3F3F3}{31.2899}         & \cellcolor[HTML]{F3F3F3}\textbf{{50.0272}}        & \cellcolor[HTML]{F3F3F3}\textbf{{76.4766}} \\
        & \cellcolor[HTML]{F3F3F3}\textbf{{(+0.5898)}}      & \cellcolor[HTML]{F3F3F3}\textbf{{(+0.0601)}}      & \cellcolor[HTML]{F3F3F3}{(+0.2439)}         & \cellcolor[HTML]{F3F3F3}\textbf{{(+0.2532)}}        & \cellcolor[HTML]{F3F3F3}\textbf{{(+0.8970)}} \\
\hline
\end{tabular}}
\label{tab:camo}
\end{table}

\begin{table}[ht]
\centering
\caption{Experimetal results of tuning \(\alpha\), \(\beta\) hyperparameters on Transformer. CIDEr (\%) is used to evaluate the performance of hyperparameter pairs.}
\resizebox{0.45\textwidth}{!}{\begin{tabular}{c|ccccc}
\hline
\diagbox{{$\mathbf{\beta}$}}{$\mathbf{\alpha}$} & \textbf{0.1}     & \textbf{0.2}     & \textbf{0.3}     & \textbf{0.4}     & \textbf{0.5}     \\
\hline
\textbf{0.1}        & 71.3749 & 71.2570  & 70.9183 & 70.7879 & 70.6420  \\ 
\textbf{0.2}        & 71.8213 & 71.9769 & 71.7687 & 71.7945 & 71.4786 \\
\textbf{0.3}        & 72.4507 & 72.2566 & 72.4534 & 72.1446  & 71.7824 \\
\textbf{0.4}        & \cellcolor[HTML]{F3F3F3}\textbf{72.7152} & 72.7141 & 72.0606 & 72.3325 & 71.5992 \\
\textbf{0.5}        & 71.3598 & 71.7182 & 71.4344 & 71.3747 & 71.1979 \\
\hline
\end{tabular}}
\end{table}

\subsection{Evaluation Metrics} We use five standard metrics to measure the performance of baselines, and our approach including BLEU \cite{papineni2002bleu}, METEOR \cite{denkowski2014meteor}, ROUGE \cite{rouge2004package}, and CIDEr \cite{vedantam2015cider}. SPICE is not used because the Vietnamese language scene graph is unavailable now.

\subsection{Implementation Details}

\textbf{Visual Features.} There are two concepts of visual features used to train captioning models: ``region-based features'' and ``grid-based features.'' ``Region-based features" indicates embedding vectors of regions of interest in input images, while ``Grid-based features" is the definition of global high-level features extracted from CNN backbones. To extract two types of visual features, we use the ResNeXt-152++ pre-trained model provided in the study \cite{gridsdefense}. In detail, \citeauthor{gridsdefense} trained a Faster R-CNN model on the VisualGenome dataset and only extracted grid features from the top of the CNN backbone. In our study, we also extract regions' features by using their further Fully Connected layers. With DIFNet, they used segmentation masks of input images as additional visual features. In this case, we use the same pre-trained model with them, UPSNet \cite{xiong2019upsnet}, to extract segmentation features. For a fair comparison, we use the same type of features reported in the original paper of baselines mentioned in Section \ref{sec:baseline}. The type of features used to train captioning models is also shown in Table \ref{tab:expresults} (\(\mathcal{R}\), \(\mathcal{G}\), \(\mathcal{L}\), \(\mathcal{S}\) denotes region-based features, grid-based features, language features, and segmentation mask, respectively). 

\textbf{Training Details.} All methods mentioned in Section \ref{sec:baseline} are Transformer-based models. Therefore, the number of encoder layers and decoder layers is equal to \(3\). Following the original Transformer architecture, there are \(8\) self-attention heads used to learn the high-level relation features. Similar to most studies, we use the two-stage training strategy. First, captioning models are optimized via Cross-Entropy loss:

\begin{equation}
\mathcal{L}_{XE}(\theta) = - \Sigma^{T}_{t=1} \log (p_{\theta}(w_t^{*}|w_{1:t-1}^{*})),
\end{equation}
where \(\theta\) denotes the learned parameters; \(w_{1:T}^{*}\) denotes the ground-truth captions.

Second, we follow \cite{scst} and apply Self-Critical Sequence Training (SCST) based on REINFORCE policy algorithm after training with Cross-Entropy loss to refine the generated captions; the loss function is now formulated as:
\begin{equation}
    \mathcal{L}_{RL}(\theta) = -E_{w_{1:T} \ p_{\theta}} \left [ r(w_{1:T}) \right ],
\end{equation}
where the reward $r(\cdot)$ denotes the CIDEr-D score.

We use Adam optimizer for training all baseline models; the difference lies in the scheduler. With the original Transformer \cite{transformer}, \(\mathcal{M}^2\) Transformer \cite{m2transformer}, ORT \cite{ort} and AoANet \cite{aoanet}, we use the original learning rate scheduling strategy introduced in \cite{transformer}, which is defined as follows:

\begin{equation}
lambda\_lr = (d_{model}) ^ {-0.5} \times \text{min}(e^{-0.5}, s \times (N_w) ^ {-0.5}),
\end{equation}
Where $d_{model}$ is the fixed number of channels of high-level features in the Transformer model, $e$ is the considering epoch. $N_w$ is the number of warmup iterations.

With DLCT \cite{luo2021dual}, RSTNet \cite{rstnet}, DIFNet \cite{difnet}, and MDSANet \cite{mdsanet}, we follow the learning rate strategy that is used in their original versions trained on MS-COCO dataset, which is defined as follows:

\begin{equation}
lambda\_lr =
    \begin{cases}
      base\_lr \times e/4 & $ \text{e \(\leq\) 3} $ \\
      base\_lr & \text{3 \(\leq\) e \(\leq\) 10}\\
      base\_lr \times 0.2 & \text{10 \(\leq\) e \(\leq\) 12} \\
      base\_lr \times 0.2^2 & \text{otherwise}
    \end{cases}
\end{equation}
where \(base\_lr\) is the initialized learning rate, which is set as 1.

With all baselines, we keep their optimization schedulers which are reported in their papers, and they are trained on our UIT-OpenViIC dataset using \(1 \times\) GPU RTX 2080Ti.'

\subsection{Quantitative Results}

\begin{figure*}[ht]
\centerline{\includegraphics[width=18cm]{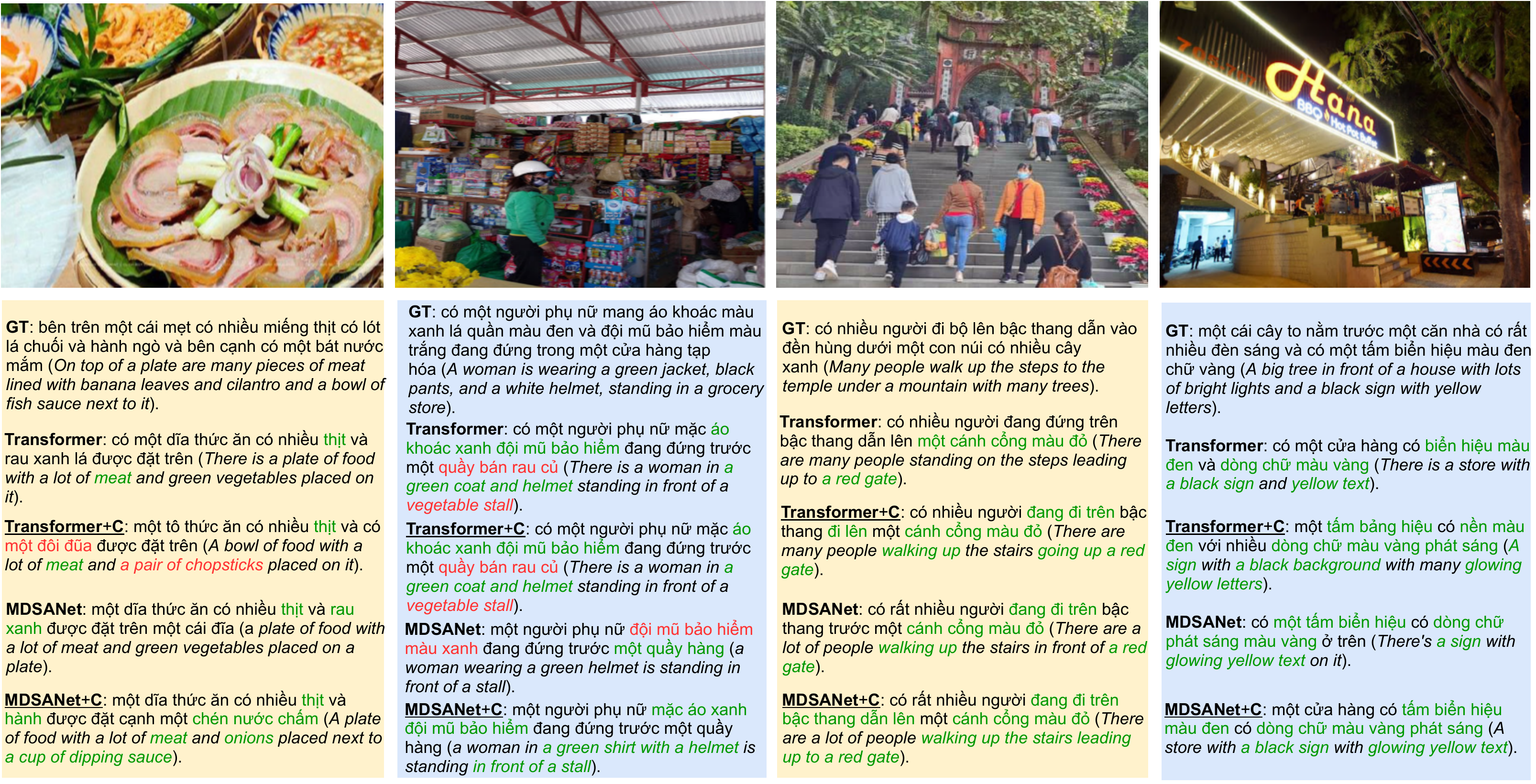}}
\caption{Examples of captions generated by our approach and the original Transformer and MDSANet models. (4 \(\times\) zooming out for the better illustration)}
\label{results-visualization}
\end{figure*}

\textbf{Main Results.} We report the experimental results of baselines in the studies \cite{ort, m2transformer, rstnet, luo2021dual, difnet, mdsanet} at Table \ref{tab:expresults}. Overall, our dataset is quite challenging for recent models to generate captions. The highest results are recorded using MDSANet \cite{mdsanet}, whose CIDEr score is 75.5796\%. It shows that the multi-branch self-attention of each head indeed gains more high-level insights from visual features and effectively guides the main model to generate quality captions. Although using multi-representation schemes, DLCT (\(\mathcal{G}\) + \(\mathcal{R}\)) and DIFNet (\(\mathcal{G}\) + \(\mathcal{S}\)) still do not achieves the expected results (72.6897\% and 74.1907\% CIDEr, respectively). The reason is that some available pre-trained models for extracting these features may fail in our UIT-OpenViIC dataset (due to complex scenes in Vietnam, which may not occur in datasets used for training available feature extractors). Consequently, it leads to incorrect information fitting into the captioning models. RSTNet adopts external language features extracted from BERT pre-trained model and guides the model to generate better non-visual words. However, Vietnamese vocabulary contains various non-visual syllables, and the appearance frequency is much less than the MS-COCO dataset, which could confuse the Language Adaptive Attention of RSTNet (72.7171\% CIDEr). Other Transformer-based models which focus on improving visual flow in encoder-decoder architecture (\(\mathcal{M}^2\), ORT) easily witness failures due to low-quality appearance features representing images in the UIT-OpenViIC dataset (66.8880\% and 68.7317\% CIDEr, respectively). Attention-on-Attention mechanism gains better results compared to \(\mathcal{M}^2\) and ORT; however, it is still modest (69.4182\% CIDEr). In each baseline, we also show its performance in MS-COCO Karpathy test split. It can be seen that there is a large margin between the highest results on the UIT-OpenViIC test set and MS-COCO Karpathy among baselines (MDSANet 75.5796\% vs. DIFNet 136.2\%). This observation raises the challenges of our UIT-OpenViIC dataset and has room to grow in the future.

\begin{figure*}[http]
\centerline{\includegraphics[width=18cm]{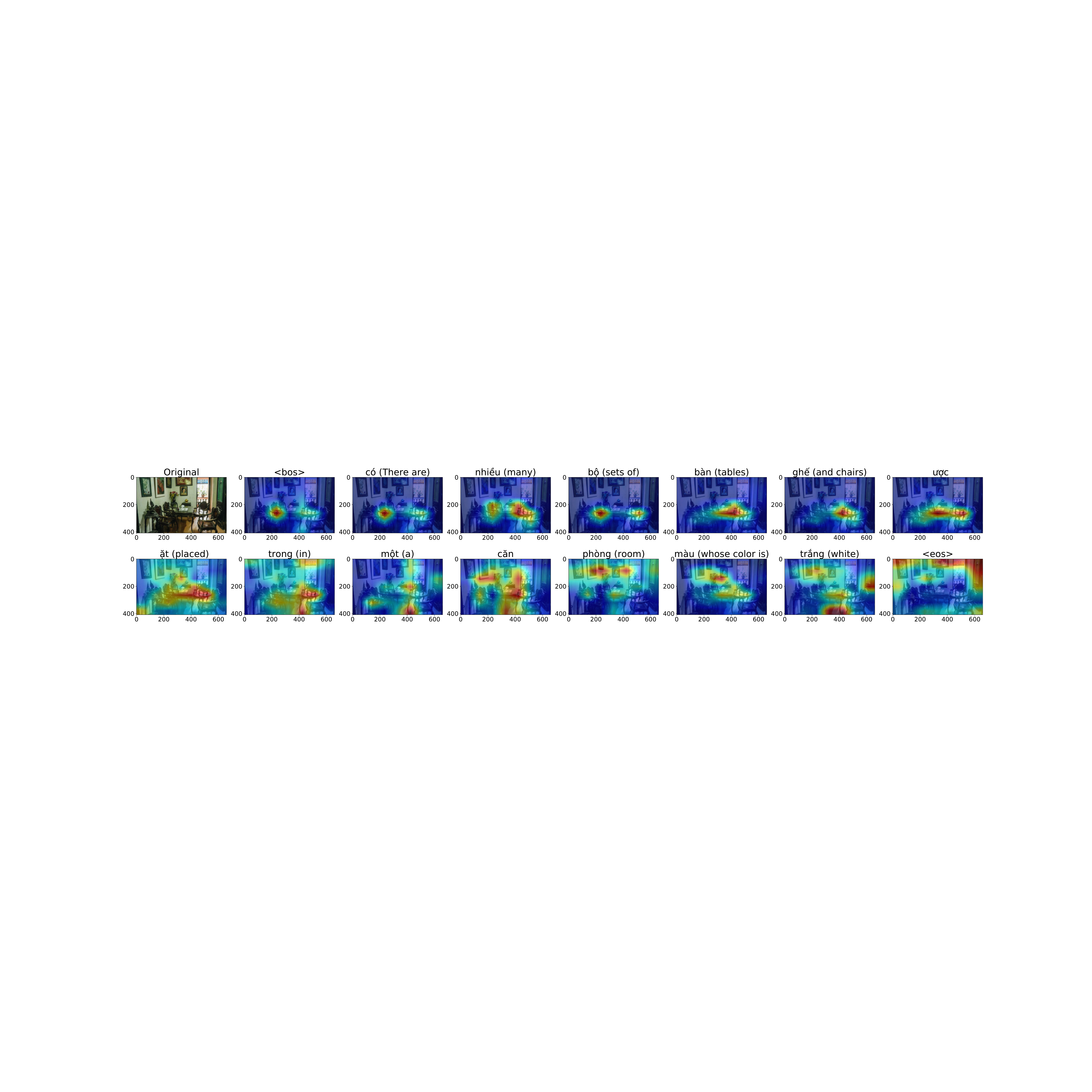}}
\caption{Visualization of attention score through decoding steps using original MDSANet. (4 \(\times\) zooming out for the better illustration)}
\label{results-attmap-mdsanet}
\end{figure*}

\begin{figure*}[http]
\centerline{\includegraphics[width=18cm]{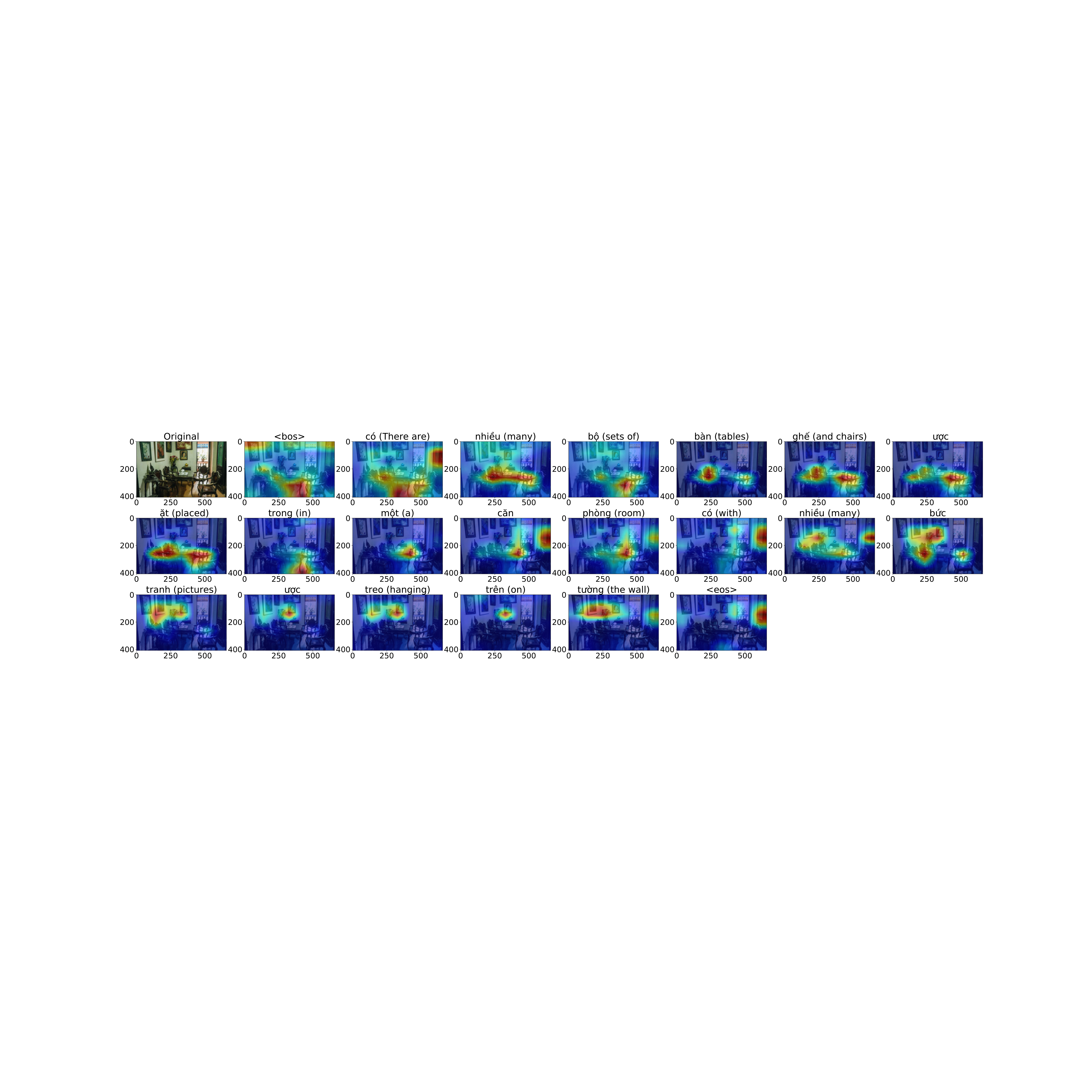}}
\caption{Visualization of attention score through decoding steps using MDSANet with proposed CAMO. (4 \(\times\) zooming out for the better illustration)}
\label{results-attmap-mdsanet-camo}
\end{figure*}

\textbf{CAMO Analysis.} Our CAMO approach is proposed to explore the encoded visual information from the previous encoder layers. We do the experiments applying our CAMO on all baselines and report the results in Table \ref{tab:camo}. Generally, our CAMO approach performs more effectively on simple captioning models than the complex ones. Indeed, CAMO substantially improves the original Transformer when obtaining 71.8213\% CIDEr, 4.9167\% higher than the original setting. ORT is a simple Transformer-based model which only uses bounding box coordinates to guide the self-attention mechanism; therefore, CAMO also helps it improve 2.4588\% CIDEr than its original. AoANet witnesses the crucial improvement (+4.5701\% CIDEr) when using CAMO. We hypothesize that the Attention-on-Attention mechanism resonates with our multi-level cross-attention, leading to better performance. The resonance of the memory-augmented encoder and CAMO approach also obtain better quality, whose CIDEr is 3.0187\% higher than the original \(\mathcal{M}2\) Transformer. RSTNet, DIFNet, and MDSANet have complex calculations that lie on Adaptive Attention, Dual Visual Flow, and Multi-branch Self-attention, respectively. Therefore, CAMO helps them sightly improve the results (RSTNet +1.4698\% CIDEr, MDSANet +0.8970\% CIDEr). We also provide the ablation study that tunning \(\alpha\) and \(\beta\) hyperparameters. \(\beta\) seems to contribute much more than \(\alpha\), when the highest results are recorded using \(\alpha=0.1\), \(\beta=0.4\).

\textbf{Regions and Grids.} For another aspect, we observe that our CAMO approach is more effective with models trained with region-based features. Indeed, images are represented by region features in the original Transformer, ORT, M2 Transformer, and AoANet, and their results achieve more significant improvement (+4.9167, +2.4588, +4.5701, +3.0187, respectively) when applying CAMO than other baselines using grid-based features. The reason is that grid-based features represent the input image globally; hence complex context is included in a grid leading to confusion for CAMO to look back.

\subsection{Qualitative Results}

\textbf{Generated Captions.} We provide the generated captions from trained models on some examples in the UIT-OpenViIC dataset in Figure \ref{results-visualization}. Qualitatively, the quality of generated captions produced via baselines applying CAMO seems higher than the originals. In the first picture, MDSANet+C recognizes there is ``a cup of dipping sauce''. In the second picture, MDSANet+\(\mathcal{C}\) accurately describes the woman wearing a green shirt with a helmet and standing in front of a stall. In the final picture, Transformer+\(\mathcal{C}\) correctly recognizes the action of ``walking'' of crowded people, and MDSANet+\(\mathcal{C}\) describes the target direction of people more accurately, which is ``leading up to a red gate''.

\textbf{Grid-based Attention Scores.} With models trained on Grid-based features, the attention scores have the shape of \(\mathrm{R}^{h \times G \times G}\). We take the mean values to unify the attention scores over \(h\) heads. First, we normalize its values to the range of \([0, 1]\) by dividing all values in attention scores with the max value. Then, we use \texttt{cv2.resize} to upsample the attention map's size to equal the resolution of the input image. Then, to make it visible, we multiply all values with 255 and create the heat map using \texttt{cv2.applyColorMap} method which the \texttt{cv2.COLORMAP\_JET} integer type. Finally, the input image and heat map are merged using \texttt{cv2.addWeighted} method, with \(\alpha = 0.3\). As illustrated in Figure \ref{results-attmap-mdsanet} and Figure \ref{results-attmap-mdsanet-camo}, the MDSANet applying CAMO seems to perform better. While the original MDSANet only looks at chairs and tables, our MDSANet+\(\mathcal{C}\) correctly observes pictures hanging on the wall, and the attention scores prove that it indeed looks at pictures of objects in the scene.

\section{Conclusion}
\label{sec:conclusion}

In this study, we introduce the UIT-OpenViIC dataset, which is the first that includes open-domain images related to the Vietnam context, and also has the most significant number of image-caption pairs among the Vietnamese dataset. Moreover, we present the Cross-attention on Multi-level Encoder Outputs (CAMO) approach to enhance the quality of high-level features produced by the Transformer Encoder. Through experiments, we prove that our method is effective and can boost the performance of existing captioning models without increasing too many learned parameters. We believe that publicly releasing our UIT-OpenViIC dataset will motivate vision-language research communities in Vietnam to be interested in the image captioning problem and continue to achieve better results on our challenging dataset. Last but not least, our dataset also makes the dataset library for image captioning in the world more varied, which consists of images in various scenes, and high-quality annotated captions for research.

\section*{Acknowledgment}
This research is funded by Vietnam National University Ho Chi Minh City (VNU-HCM) under grant number C2023-26-11.

\printbibliography

@String(CVPR= {IEEE Conf. Comput. Vis. Pattern Recog.})

@String(ICCV= {Int. Conf. Comput. Vis.})

@String(ECCV= {Eur. Conf. Comput. Vis.})

@String(NIPS= {Adv. Neural Inform. Process. Syst.})

@String(AAAI = {AAAI})

@String(CVPR  = {CVPR})

@String(ICCV  = {ICCV})

@String(ECCV  = {ECCV})

@String(NIPS  = {NeurIPS})

@article{chen2015microsoft,
  title={Microsoft coco captions: Data collection and evaluation server},
  author={Chen, Xinlei and Fang, Hao and Lin, Tsung-Yi and Vedantam, Ramakrishna and Gupta, Saurabh and Doll{\'a}r, Piotr and Zitnick, C Lawrence},
  journal={arXiv preprint arXiv:1504.00325},
  year={2015}
}

@INPROCEEDINGS{7298932,
  author={Karpathy, Andrej and Fei-Fei, Li},
  booktitle={2015 IEEE Conference on Computer Vision and Pattern Recognition (CVPR)}, 
  title={Deep visual-semantic alignments for generating image descriptions}, 
  year={2015},
  volume={},
  number={},
  pages={3128-3137},
  doi={10.1109/CVPR.2015.7298932}}

@inproceedings{le2021vlsp,
title = {VLSP 2021 - VieCap4H Challenge: Automatic Image Caption Generation for Healthcare Domain in Vietnamese},
author = {Le, Thao Minh and Dang, Long Hoang and Nguyen, Thanh-Son and Nguyen, Thi Minh Huyen and Vu, Xuan-Son},
booktitle = {Proceedings of the 8th International Workshop on Vietnamese Language and Speech Processing},
month = {12}, year = {2021}, address = {Ho Chi Minh, Vietnam},
publisher = {VNU Journal of Science: Computer Science and Communication Engineering}}

@InProceedings{uitviic,
author="Lam, Quan Hoang
and Le, Quang Duy
and Nguyen, Van Kiet
and Nguyen, Ngan Luu-Thuy",
editor="Nguyen, Ngoc Thanh
and Hoang, Bao Hung
and Huynh, Cong Phap
and Hwang, Dosam
and Trawi{\'{n}}ski, Bogdan
and Vossen, Gottfried",
title="UIT-ViIC: A Dataset for the First Evaluation on Vietnamese Image Captioning",
booktitle="Computational Collective Intelligence",
year="2020",
publisher="Springer International Publishing",
address="Cham",
pages="730--742",
isbn="978-3-030-63007-2"
}

@ARTICLE{crowdedcaption,
  author={Wang, Lanxiao and Li, Hongliang and Hu, Wenzhe and Zhang, Xiaoliang and Qiu, Heqian and Meng, Fanma and Wu, Qingbo},
  journal={IEEE Transactions on Multimedia}, 
  title={What Happens in Crowd Scenes: A New Dataset about Crowd Scenes for Image Captioning}, 
  year={2022},
  volume={},
  number={},
  pages={1-13},
  doi={10.1109/TMM.2022.3192729}}

@INPROCEEDINGS{flickr30K,
author={Plummer, Bryan A. and Wang, Liwei and Cervantes, Chris M. and Caicedo, Juan C. and Hockenmaier, Julia and Lazebnik, Svetlana},
booktitle={2015 IEEE International Conference on Computer Vision (ICCV)}, 
title={Flickr30k Entities: Collecting Region-to-Phrase Correspondences for Richer Image-to-Sentence Models}, 
year={2015},
volume={},
number={},
pages={2641-2649},
doi={10.1109/ICCV.2015.303}}

@article{krishna2017visual,
  title={Visual genome: Connecting language and vision using crowdsourced dense image annotations},
  author={Krishna, Ranjay and Zhu, Yuke and Groth, Oliver and Johnson, Justin and Hata, Kenji and Kravitz, Joshua and Chen, Stephanie and Kalantidis, Yannis and Li, Li-Jia and Shamma, David A and others},
  journal={International journal of computer vision},
  volume={123},
  number={1},
  pages={32--73},
  year={2017},
  publisher={Springer}
}

@InProceedings{textcaps,
author="Sidorov, Oleksii
and Hu, Ronghang
and Rohrbach, Marcus
and Singh, Amanpreet",
editor="Vedaldi, Andrea
and Bischof, Horst
and Brox, Thomas
and Frahm, Jan-Michael",
title="TextCaps: A Dataset for Image Captioning with Reading Comprehension",
booktitle="Computer Vision -- ECCV 2020",
year="2020",
publisher="Springer International Publishing",
address="Cham",
pages="742--758",
isbn="978-3-030-58536-5"
}

@inproceedings{biten2019good,
  title={Good news, everyone! context driven entity-aware captioning for news images},
  author={Biten, Ali Furkan and Gomez, Lluis and Rusinol, Mar{\c{c}}al and Karatzas, Dimosthenis},
  booktitle={Proceedings of the IEEE/CVF Conference on Computer Vision and Pattern Recognition},
  pages={12466--12475},
  year={2019}
}

@inproceedings{fasterrcnn,
author = {Ren, Shaoqing and He, Kaiming and Girshick, Ross and Sun, Jian},
title = {Faster R-CNN: Towards Real-Time Object Detection with Region Proposal Networks},
year = {2015},
publisher = {MIT Press},
address = {Cambridge, MA, USA},
booktitle = {Proceedings of the 28th International Conference on Neural Information Processing Systems - Volume 1},
pages = {91–99},
numpages = {9},
location = {Montreal, Canada},
series = {NIPS'15}
}

@inbook{ort,
author = {Herdade, Simao and Kappeler, Armin and Boakye, Kofi and Soares, Joao},
title = {Image Captioning: Transforming Objects into Words},
year = {2019},
publisher = {Curran Associates Inc.},
address = {Red Hook, NY, USA},
booktitle = {Proceedings of the 33rd International Conference on Neural Information Processing Systems},
articleno = {999},
numpages = {11}
}

@inproceedings{transformer,
author = {Vaswani, Ashish and Shazeer, Noam and Parmar, Niki and Uszkoreit, Jakob and Jones, Llion and Gomez, Aidan N. and Kaiser, \L{}ukasz and Polosukhin, Illia},
title = {Attention is All You Need},
year = {2017},
isbn = {9781510860964},
publisher = {Curran Associates Inc.},
address = {Red Hook, NY, USA},
booktitle = {Proceedings of the 31st International Conference on Neural Information Processing Systems},
pages = {6000–6010},
numpages = {11},
location = {Long Beach, California, USA},
series = {NIPS'17}
}

@INPROCEEDINGS{m2transformer,  author={Cornia, Marcella and Stefanini, Matteo and Baraldi, Lorenzo and Cucchiara, Rita},  booktitle={2020 IEEE/CVF Conference on Computer Vision and Pattern Recognition (CVPR)},   title={Meshed-Memory Transformer for Image Captioning},   year={2020},  volume={},  number={},  pages={10575-10584},  doi={10.1109/CVPR42600.2020.01059}}

@INPROCEEDINGS{aoanet,  author={Huang, Lun and Wang, Wenmin and Chen, Jie and Wei, Xiao-Yong},  booktitle={2019 IEEE/CVF International Conference on Computer Vision (ICCV)},   title={Attention on Attention for Image Captioning},   year={2019},  volume={},  number={},  pages={4633-4642},  doi={10.1109/ICCV.2019.00473}}

@INPROCEEDINGS{rstnet,  author={Zhang, Xuying and Sun, Xiaoshuai and Luo, Yunpeng and Ji, Jiayi and Zhou, Yiyi and Wu, Yongjian and Huang, Feiyue and Ji, Rongrong},  booktitle={2021 IEEE/CVF Conference on Computer Vision and Pattern Recognition (CVPR)},   title={RSTNet: Captioning with Adaptive Attention on Visual and Non-Visual Words},   year={2021},  volume={},  number={},  pages={15460-15469},  doi={10.1109/CVPR46437.2021.01521}}

@inproceedings{luo2021dual,
  title={Dual-level collaborative transformer for image captioning},
  author={Luo, Yunpeng and Ji, Jiayi and Sun, Xiaoshuai and Cao, Liujuan and Wu, Yongjian and Huang, Feiyue and Lin, Chia-Wen and Ji, Rongrong},
  booktitle={Proceedings of the AAAI Conference on Artificial Intelligence},
  volume={35},
  number={3},
  pages={2286--2293},
  year={2021}
}

@INPROCEEDINGS{difnet,
  author={Wu, Mingrui and Zhang, Xuying and Sun, Xiaoshuai and Zhou, Yiyi and Chen, Chao and Gu, Jiaxin and Sun, Xing and Ji, Rongrong},
  booktitle={2022 IEEE/CVF Conference on Computer Vision and Pattern Recognition (CVPR)}, 
  title={DIFNet: Boosting Visual Information Flow for Image Captioning}, 
  year={2022},
  volume={},
  number={},
  pages={17999-18008},
  doi={10.1109/CVPR52688.2022.01749}}

@ARTICLE{mdsanet,  author={Ji, Jiayi and Huang, Xiaoyang and Sun, Xiaoshuai and Zhou, Yiyi and Luo, Gen and Cao, Liujuan and Liu, Jianzhuang and Shao, Ling and Ji, Rongrong},  journal={IEEE Transactions on Multimedia},   title={Multi-Branch Distance-Sensitive Self-Attention Network for Image Captioning},   year={2022},  volume={},  number={},  pages={1-1},  doi={10.1109/TMM.2022.3169061}}

@inproceedings{anderson2018bottom,
  title={Bottom-up and top-down attention for image captioning and visual question answering},
  author={Anderson, Peter and He, Xiaodong and Buehler, Chris and Teney, Damien and Johnson, Mark and Gould, Stephen and Zhang, Lei},
  booktitle={Proceedings of the IEEE conference on computer vision and pattern recognition},
  pages={6077--6086},
  year={2018}
}

@article{nagrani2021attention,
  title={Attention bottlenecks for multimodal fusion},
  author={Nagrani, Arsha and Yang, Shan and Arnab, Anurag and Jansen, Aren and Schmid, Cordelia and Sun, Chen},
  journal={Advances in Neural Information Processing Systems},
  volume={34},
  pages={14200--14213},
  year={2021}
}

@article{hossain2019survey,
author = {Hossain, MD. Zakir and Sohel, Ferdous and Shiratuddin, Mohd Fairuz and Laga, Hamid},
title = {A Comprehensive Survey of Deep Learning for Image Captioning},
year = {2019},
issue_date = {November 2019},
publisher = {Association for Computing Machinery},
address = {New York, NY, USA},
volume = {51},
number = {6},
issn = {0360-0300},
url = {https://doi.org/10.1145/3295748},
doi = {10.1145/3295748},
journal = {ACM Comput. Surv.},
month = {2},
articleno = {118},
numpages = {36}
}

@INPROCEEDINGS{scst,  author={Rennie, Steven J. and Marcheret, Etienne and Mroueh, Youssef and Ross, Jerret and Goel, Vaibhava},  booktitle={2017 IEEE Conference on Computer Vision and Pattern Recognition (CVPR)},   title={Self-Critical Sequence Training for Image Captioning},   year={2017},  volume={},  number={},  pages={1179-1195},  doi={10.1109/CVPR.2017.131}}

@article{vu2018vncorenlp,
  title={VnCoreNLP: A Vietnamese natural language processing toolkit},
  author={Vu, Thanh and Nguyen, Dat Quoc and Nguyen, Dai Quoc and Dras, Mark and Johnson, Mark},
  journal={arXiv preprint arXiv:1801.01331},
  year={2018}
}

@inproceedings{papineni2002bleu,
  title={Bleu: a method for automatic evaluation of machine translation},
  author={Papineni, Kishore and Roukos, Salim and Ward, Todd and Zhu, Wei-Jing},
  booktitle={Proceedings of the 40th annual meeting of the Association for Computational Linguistics},
  pages={311--318},
  year={2002}
}

@inproceedings{denkowski2014meteor,
  title={Meteor universal: Language specific translation evaluation for any target language},
  author={Denkowski, Michael and Lavie, Alon},
  booktitle={Proceedings of the ninth workshop on statistical machine translation},
  pages={376--380},
  year={2014}
}

@inproceedings{rouge2004package,
  title={A package for automatic evaluation of summaries},
  author={ROUGE, Lin CY},
  booktitle={Proceedings of Workshop on Text Summarization of ACL, Spain},
  year={2004}
}

@inproceedings{vedantam2015cider,
  title={Cider: Consensus-based image description evaluation},
  author={Vedantam, Ramakrishna and Lawrence Zitnick, C and Parikh, Devi},
  booktitle={Proceedings of the IEEE conference on computer vision and pattern recognition},
  pages={4566--4575},
  year={2015}
}

@INPROCEEDINGS{gridsdefense,  author={Jiang, Huaizu and Misra, Ishan and Rohrbach, Marcus and Learned-Miller, Erik and Chen, Xinlei},  booktitle={2020 IEEE/CVF Conference on Computer Vision and Pattern Recognition (CVPR)},   title={In Defense of Grid Features for Visual Question Answering},   year={2020},  volume={},  number={},  pages={10264-10273},  doi={10.1109/CVPR42600.2020.01028}}

@inproceedings{xiong2019upsnet,
  title={Upsnet: A unified panoptic segmentation network},
  author={Xiong, Yuwen and Liao, Renjie and Zhao, Hengshuang and Hu, Rui and Bai, Min and Yumer, Ersin and Urtasun, Raquel},
  booktitle={Proceedings of the IEEE/CVF Conference on Computer Vision and Pattern Recognition},
  pages={8818--8826},
  year={2019}
}
\end{document}